\documentclass[letterpaper, 10 pt, conference]{ieeeconf}  % Comment this line out if you need a4paper

\IEEEoverridecommandlockouts                              % This command is only needed if 
\usepackage[usenames,dvipsnames]{xcolor}

\newtheorem{problem}{Problem}
\usepackage{amsmath}
\usepackage{graphicx}
\usepackage{caption}
\usepackage{subcaption}
\usepackage{balance}

\title{\LARGE \bf
Guided Learning from Demonstration for Robust Transferability
}

\author{Fouad Sukkar*, Victor Hernandez Moreno*, Teresa Vidal-Calleja and Jochen Deuse% <-this % stops a space
\thanks{*These authors are co-first authors and contributed equally to this work.\newline
This research is partially supported by the University of Technology Sydney's President and International Research Scholarships and the Industrial Transformation Training Centre (ITTC) for Collaborative Robotics in Advanced Manufacturing (also known as the Australian Cobotics Centre) funded by ARC (Project ID: IC200100001).}%
\thanks{Authors are with the UTS Robotics Institute and the Centre for Advanced Manufacturing, School for Mechanical and Mechatronic Engineering, University of Technology Sydney, 2007, Ultimo, NSW, Australia and the Australian Cobotics Centre (ITTC for Collaborative Robotics in Advanced Manufacturing)
{\tt\footnotesize \{Fouad.Sukkar@,Victor.HernandezMoreno@student., Teresa.VidalCalleja@,Jochen.Deuse@\}uts.edu.au}.}% <-this % stops a space
}

\begin{document}

\maketitle
\thispagestyle{empty}
\pagestyle{empty}

%%%%%%%%%%%%%%%%%%%%%%%%%%%%%%%%%%%%%%%%%%%%%%%%%%%%%%%%%%%%%%%%%%%%%%%%%%%%%%%%
\begin{abstract}
Learning from demonstration (LfD) has the potential to greatly increase the applicability of robotic manipulators in modern industrial applications. Recent progress in LfD methods have put more emphasis in learning robustness than in guiding the demonstration itself in order to improve robustness. The latter is particularly important to consider when the target system reproducing the motion is structurally different to the demonstration system, as some demonstrated motions may not be reproducible. In light of this, this paper introduces a new guided learning from demonstration paradigm where an interactive graphical user interface (GUI) guides the user during demonstration, preventing them from demonstrating non-reproducible motions. The key aspect of our approach is determining the space of reproducible motions based on a motion planning framework which finds regions in the task space where trajectories are guaranteed to be of bounded length. We evaluate our method on two different setups with a six-degree-of-freedom (DOF) UR5 as the target system. First our method is validated using a seven-DOF Sawyer as the demonstration system. Then an extensive user study is carried out where several participants are asked to demonstrate, with and without guidance, a mock weld task using a hand held tool tracked by a VICON system. With guidance users were able to always carry out the task successfully in comparison to only 44\% of the time without guidance.

\end{abstract}

%%%%%%%%%%%%%%%%%%%%%%%%%%%%%%%%%%%%%%%%%%%%%%%%%%%%%%%%%%%%%%%%%%%%%%%%%%%%%%%%
\section{INTRODUCTION}

In modern industrial robotic manipulator applications there is a desire for greater autonomy through adaptability to novel tasks without requiring time consuming and costly reprogramming. Unlocking this potential would present the factories of tomorrow with the opportunity to shift from mass production towards mass customisation~\cite{Petersen2016}. Learning from demonstration (LfD) methods are a promising direction for achieving this autonomy~\cite{2016HandbookRobotics}. LfD enables non-robot experts to intuitively program robots that reproduce motions with high precision. 
However, often LfD methods limit this potential by restricting the demonstration to occur on the system reproducing the motion, for example via kinaesthetic teaching or teleoperation in order to guarantee faithful motion reproduction~\cite{Rozo2013, Rana2017, Zhu2018, Savarimuthu2018, Gao2019, Kyrarini2019, Liu2020a, Su2021}.

\begin{figure}[t]
    \centering
    \includegraphics[width=1.0\columnwidth]{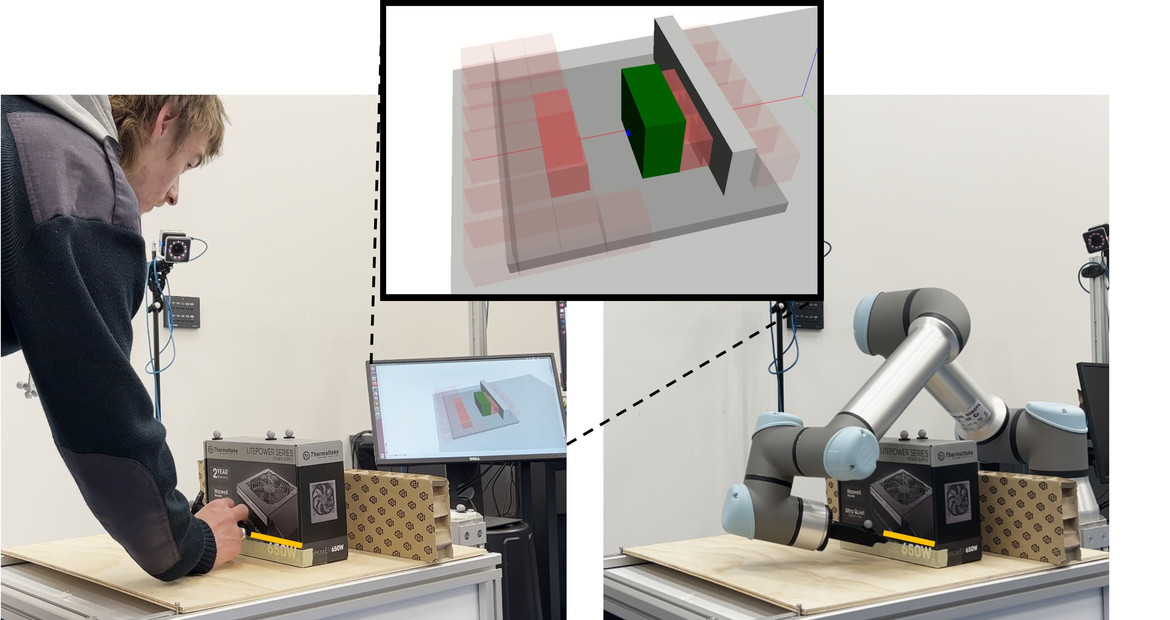}\label{fig:setup_rep_real}
    \caption{Our proposed guided learning from demonstration approach with interactive GUI (inset) showing regions of reproducible motions to the demonstrator based on the kinematics of the target robot system. The demonstrator shown here is a non-expert and is carrying out the demonstration without the physical robot present.}
    \label{fig:frontpage}
\end{figure}

We are interested in removing this restriction and allowing the demonstration and target systems to be different.
Motivating use-cases include the following: demonstrating tasks naturally by tracking human motion directly or with an unconstrained tool detached from the robot, carrying out demonstrations in parallel while the target robot executes other tasks, and demonstrating on a smaller/safer robot in order to carry out dangerous or labour intensive tasks with a more powerful robot.

The challenge in this scenario is that there is no longer a direct mapping between the demonstration and target systems due to their differing kinematic structures. A common strategy is to encode the demonstration as the robot's end-effector trajectory. Doing so provides a common representation of the motion that can be interpreted by the target system~\cite{Ravichandar2020, Vakanski2017}.
However, in these settings the kinematic discrepancy is often ignored or worked around by utilising a redundant robot for motion reproduction~\cite{Lioutikov2015, Lauretti2019, Jaquier2021, Abu-Dakka2020}.
These robots have more degrees of freedom (DOF) than the dimension of the task space~\cite{lynch2017modern} and can utilise null-space control methods in order to satisfy kinematic constraints, such as avoiding obstacles, singularities and joint limits, whilst maintaining faithful reproduction of the demonstrated end-effector motion~\cite{2016HandbookRobotics}. 

In many applications a six-DOF arm may be preferable over a seven-DOF arm due to being relatively lower cost, exhibiting higher payload-to-weight ratio and having simpler kinematics for control and planning. However, these arms cannot exploit null-space control~\cite{lynch2017modern} due to their kinematic structure and as a consequence may require large joint changes between intermediate points of a demonstrated trajectory in order to satisfy kinematic constraints.  

Hence the only option is to artificially restrict the motions of the demonstrator. While there exists work along this line through the form of graphical user interfaces (GUI)~\cite{Mollard2015} and augmented reality~\cite{Luebbers2019, Diehl2020, Luebbers2021}, these methods usually consider only verifying, adjusting or adding loosely defined workspace and task constraints which do not directly consider the kinematic constraints of the target system.

In this paper we formulate a new guided learning from demonstration problem which explicitly considers these kinematic constraints and prevents non-reproducible motions from being demonstrated.  
To compute the space of reproducible motions we leverage an existing motion planning framework called Hausdorff approximation planner (HAP)~\cite{sukkar2022motion} which finds regions in the task space where trajectories are guaranteed to be of bounded length. A novel interactive graphical user interface (GUI) then visualises this space to the user at demonstration time. Our GUI is intuitive and requires no expert knowledge about the target system from the demonstrator (see Fig.~\ref{fig:frontpage}). For encoding and reproducing demonstrated motions, we use Cartesian space Dynamic Movement Primitives (CDMPs)~\cite{Koutras2019} which facilitate a common representation space. 

We evaluate our method in a workbench environment on two sets of experiments with a six-DOF UR5 as the target system. First our method is validated using a seven-DOF Sawyer as the demonstration system. Then an extensive user study is carried out where several participants are asked to demonstrate a mock weld task using a hand held tool tracked by a VICON system. Users carry out the task with and without the interactive GUI and results show that with guidance from the GUI users had~100\% success rate in carrying out the task versus~44\% without and were on average~30\% more confident in carrying out the task.

The significance of our framework is that it enables non-expert demonstrators to confidently carry out feasible demonstrations without the presence of the physical target system and enables a greater range of possible demonstration and target system setups. The key contributions in this paper are as follows: (1) A new guided learning from demonstration problem formulation for ensuring robust motion transfer onto a target system. (2) A novel approach to tackle this problem via a method for computing regions of reproducible motions and an interactive GUI for displaying these regions to the demonstrator.

\section{PROBLEM FORMULATION}
In this section, we formulate the problem of ensuring robust motion transfer from a demonstration system to a target system with differing kinematic structures.
First we describe the problem setup including assumptions about the demonstration system, target system, environment and operational scenario. Then we formulate and state the problem of finding a region of reproducible motions for ensuring robust transferability.

\subsection{Problem Setup}
We consider a non-trivial LfD scenario where we wish to reproduce a demonstrated motion from one system onto another target system with different kinematic structure which may, for example, have less degrees of freedom (DOF).
We assume access to the target system's kinematic model and some approximate knowledge of the operational space of the demonstrator. The environment consists of static elements such as fixtures and equipment, known a-priori, and dynamic objects that can be added and removed online.

We define the demonstration space as $\mathcal{C}_{dem}$ and the demonstration system as $\mathcal{S}_{dem}$.
The desired motion is captured during the demonstration as a set of discrete states over time $\lambda_{dem}=\{t_i,x_i\}\big|^T_{i=0}$ with $x_i \in \mathcal{C}_{dem}$.
A \emph{task model}~\cite{Vakanski2017} is trained on $\lambda_{dem}$ to facilitate generalisation capabilities, such as temporal scaling, during motion reproduction.
The trained task model generates a 
trajectory $\lambda_{tar}$ which is used for reproduction on the target system $\mathcal{S}_{tar}$ within its reachable space $\mathcal{C}_{tar}$. We assume $\mathcal{C}_{dem}$ and $\mathcal{C}_{tar}$ are in a \emph{common representation space} which we choose to be $SE(3)$ since motions within this space can be interpreted by both systems. We further assume that $\mathcal{C}_{dem} \cap \mathcal{C}_{tar} \neq \emptyset$.
In order to realise the reproduced motion on the target system a \emph{motion generator} interprets $\lambda_{tar}$ and generates executable controls, $\Pi: \lambda_{tar} \mapsto \pi_{tar}$, where $\pi_{tar}$ is a sequence of control actions. 

\subsection{Region of Reproducible Motions}

In general, LfD methods impose little to no assumptions on the structure of $\mathcal{C}_{tar}$. However, for $\mathcal{S}_{dem}$ and $\mathcal{S}_{tar}$ with differing kinematic structures this is problematic since $\mathcal{C}_{tar}$ may not entirely cover $\mathcal{C}_{dem}$, that is $(\mathcal{C}_{dem} \cap \mathcal{C}_{tar}) \subset \mathcal{C}_{dem}$. Furthermore, not all $\lambda_{dem}$ are reproducible on $\mathcal{S}_{tar}$.

Thus to improve the robustness of the reproduction process we wish to constrain $\mathcal{C}_{dem}$ such that it is contained within $\mathcal{C}_{tar}$ and any $\lambda_{dem}$ within this constrained space is reproducible on $\mathcal{S}_{tar}$.
More concretely, we define a reproducible motion to be a $\lambda_{dem}$ such that the resulting $\lambda_{tar}$ when mapped through $\Pi$ results in a short, smooth path in $C_{tar}$ and is collision-free.

\begin{problem}[Region of reproducible motions] \label{prob:lfd_transferability}
Find a region $\mathcal{R} \subset \mathcal{C}_{dem}$ such that it is contained within $C_{tar}$ and any $\lambda_{dem}$ through this region is a reproducible motion. In the case that multiple distinct regions exist, find $\mathcal{R}*$ that maximises coverage of the anticipated operational space.
\end{problem}

\section{GUIDED LEARNING FROM DEMONSTRATION}
In this section we describe our approach to finding $\mathcal{R}*$ and then how we display this visually in order to guide the demonstrator. Additionally we outline our chosen task model for encoding and reproducing the demonstrated motions. 
Finally, we explain the motion generator which is a non-local control strategy where trajectories generated from the task model are converted into appropriate controls for the target system.

\subsection{Finding Regions of Reproducible Motions}

\begin{figure}[t]
    \centering
    \subfloat[Example workbench environment with wall obstacle]{
    \includegraphics[width=0.64\columnwidth]{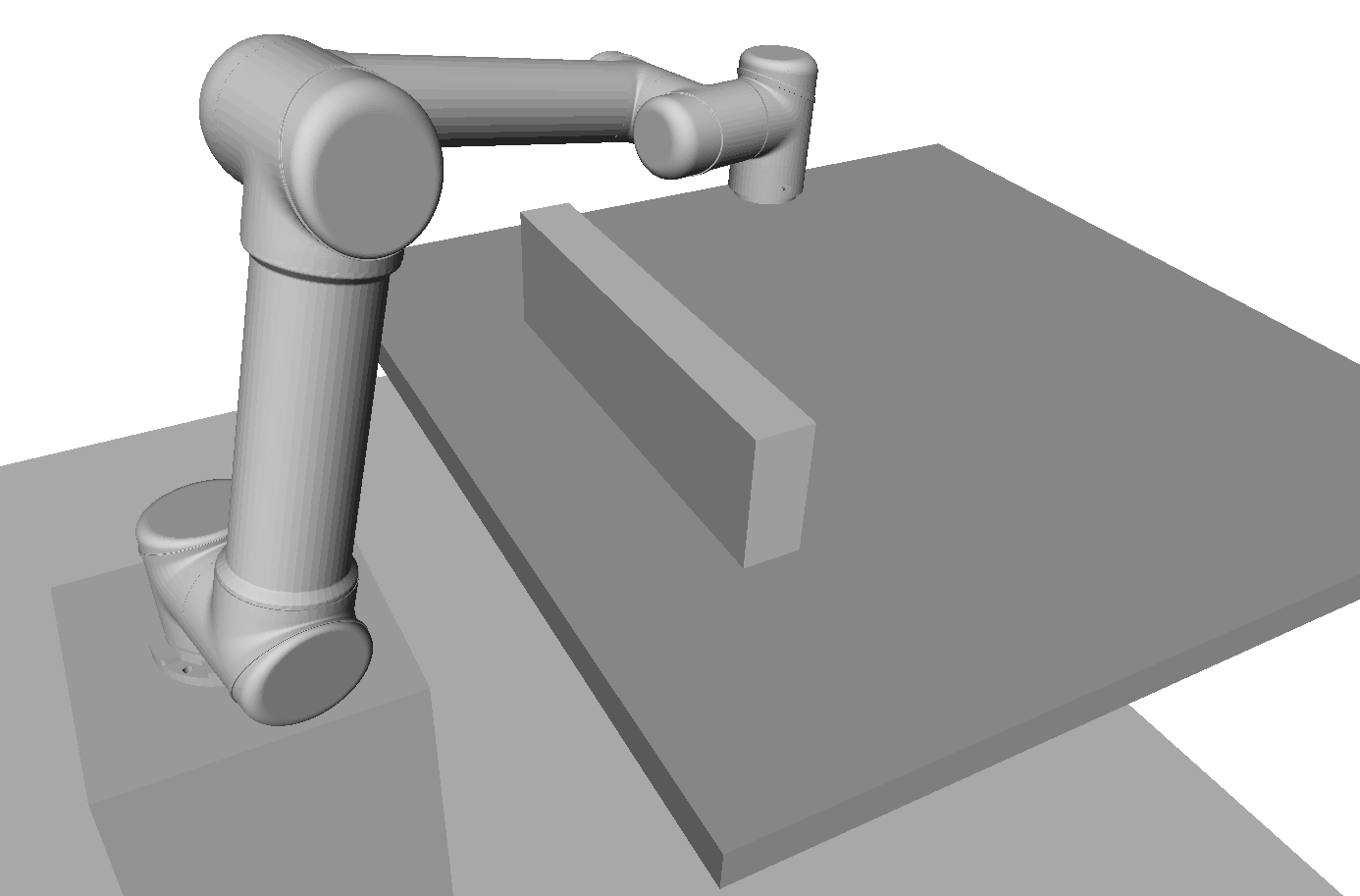}\label{fig:lfd_scenario}}
    \hfil
    \subfloat[Discrete representation of the region of reproducible motions found by HAP]{
    \includegraphics[width=0.64\columnwidth]{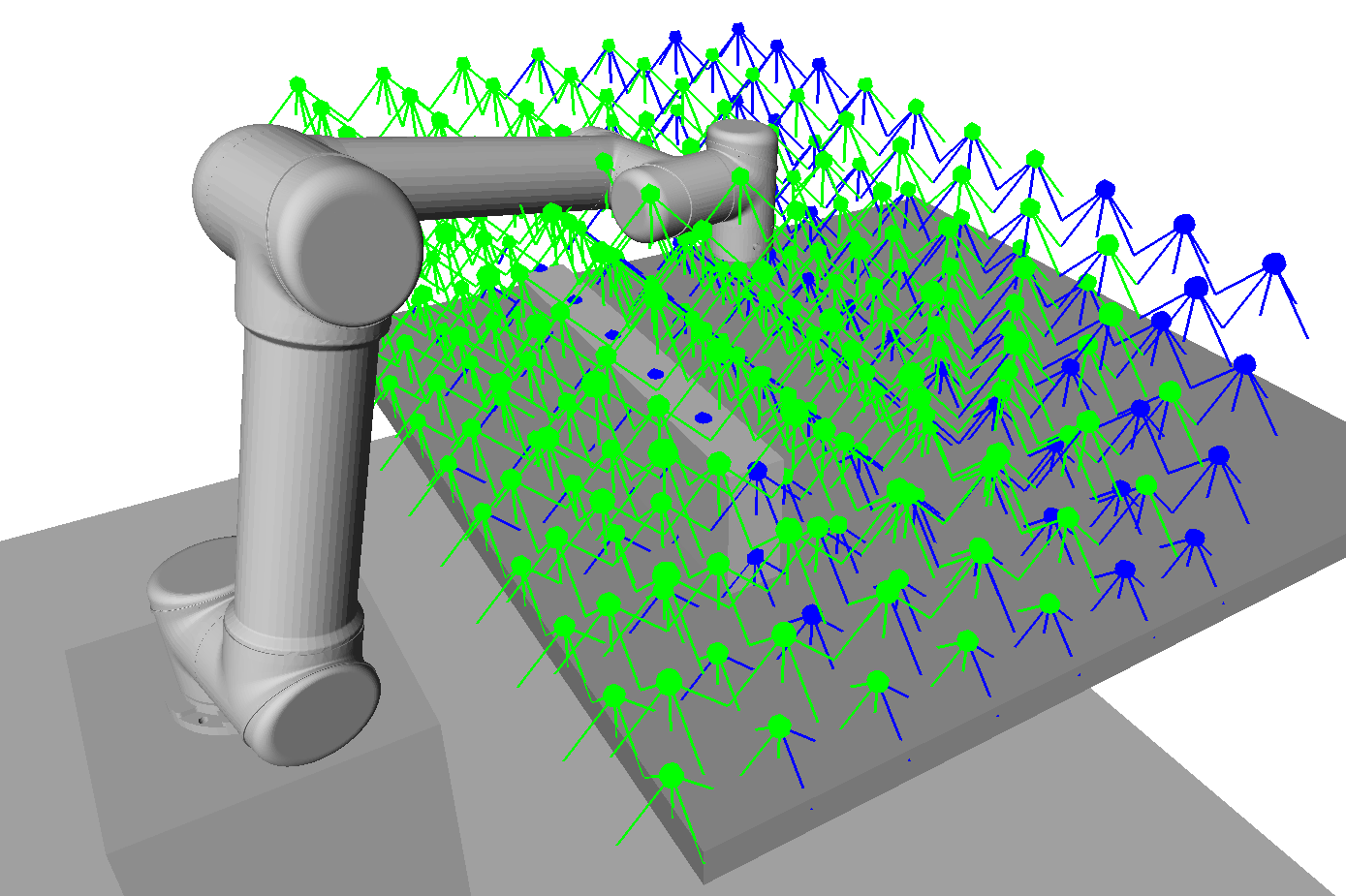}\label{fig:hap_subspace}}
    \caption{Example environment and operational space. (\protect\subref{fig:lfd_scenario}) shows a workbench environment with wall obstacle and six-DOF UR5 robot model. (\protect\subref{fig:hap_subspace}) shows a discrete task space design covering the anticipated operational space of the demonstrator where green poses indicate they are within the region of reproducible motions found by HAP and blue poses are outside this region. }
    \label{fig:example_lfd_scenario}
\end{figure}

Here we describe the method for solving Problem~\ref{prob:lfd_transferability}. In order to compute $\mathcal{R}*$ we utilise an existing robotic manipulator planning framework called Hausdorff approximation planner (HAP)~\cite{sukkar2022motion}. HAP operates on a user defined task space, in our case $\mathcal{C}_{tar} \subset SE(3)$, and divides it into one or more subspaces such that the path between any two points close in a particular subspace maps to a short, smooth and collision-free configuration space trajectory. 

The key idea is that continuous task-space trajectories can be constructed through these subspaces and the resulting configuration space trajectory will have a bounding relation in terms of trajectory distance. Such a mapping between metric spaces is called an $\epsilon$-Gromov-Hausdorff approximation ($\epsilon$-GHA). Thus, any $\lambda_{dem}$ that is contained within such a subspace in $\mathcal{C}_{tar}$ will be satisfy our condition for being reproducible.

HAP generates these maps in a pre-processing step. The procedure for generating these maps is as follows. First, HAP is given as input a discretised task space which approximately models the operational space, a robot model and an environment model. An example of each are shown in Fig.~\ref{fig:example_lfd_scenario}. An undirected graph is then constructed over the discrete task space where poses within a ball radius of each other are connected by an edge. An optimisation routine then iteratively maps unique configurations to each pose while ensuring that neighbouring poses, i.e. those connected by an edge, are within some specified bounded distance. Kinematic constraints are additionally considered, such as avoiding collision and remaining within joint limits.

The $\epsilon$-GHA that minimises the sum of all path costs through the graph is kept. This process can be repeated multiple times with a penalty added to paths that pass through previously mapped poses in order to find multiple distinct subspaces. We choose the largest covering subspace found by HAP to be $\mathcal{R}*$.

In addition, we extend the capability of HAP by performing an update to $\mathcal{R}*$ when dynamic objects are added to the environment. We perform this update by iterating over all the mapped configurations and check that they are still collision-free and if not remove them from $\mathcal{R}*$.

\subsection{Interactive Visual Guidance}

\begin{figure}[b]
    \centering
    \includegraphics[width=0.95\columnwidth]{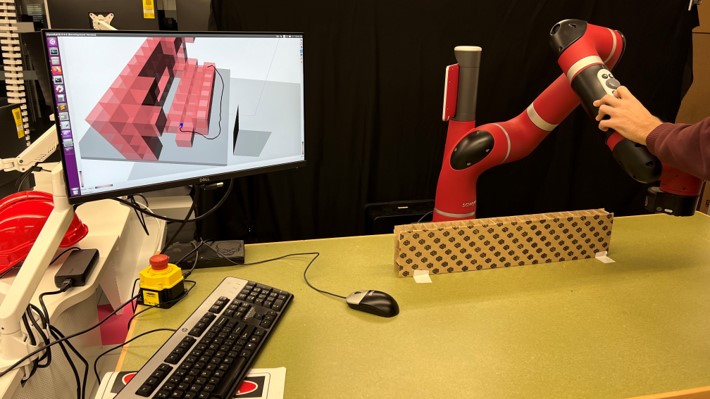}\label{fig:setup_rep_real}
    \caption{Demonstration system setup for validation experiments consisting of a seven-DOF Sawyer robot, workbench environment with a wall obstacle and interactive GUI shown on a monitor.}
    \label{fig:exp_sawyer}
\end{figure}

Given the found $\mathcal{R}*$ we wish to utilise this representation to guide demonstrations such that $\lambda_{dem} \subset \mathcal{R}*$. In order to do so, we design a graphical user interface (GUI) which displays to the user where the pose of $\mathcal{S}_{dem}$ is within $\mathcal{R}*$. Note that this GUI should be intuitive enough for a non-expert user to interpret and make effective use of without any knowledge about the underlying kinematics of the target system.

The main idea of the GUI is to block the demonstrator from moving $\mathcal{S}_{dem}$ into regions of the task space that are outside of $\mathcal{R}*$, which we will refer to as $\bar{\mathcal{R}*}$. The representation of $\mathcal{R}*$ output by HAP is a discrete set of poses in $SE(3)$. One option is to display a dense grid of arrows at each of these poses, similar to Fig.~\ref{fig:hap_subspace}; however, this would be too cluttered and confusing for the user to follow. Instead we compact six dimensional space - translation and orientation - into three dimensions by blocking the regions in $\bar{\mathcal{R}*}$ dynamically when necessary. We achieve this by only displaying red voxels at poses that are within a similar orientation to the current pose of $\mathcal{S}_{dem}$. We compute this similarity by taking the dot product between the forward pose axes of $\mathcal{S}_{dem}$ and $\bar{\mathcal{R}*}$ and only display voxels corresponding to poses within a given similarity threshold. In order to track where $\mathcal{S}_{dem}$ is in the environment, we display an arrow showing its position and forward pose axis during demonstration.

\begin{figure*}[!ht]
     \centering
     \begin{subfigure}[b]{0.24\textwidth}
         \centering
         \includegraphics[width=\textwidth]{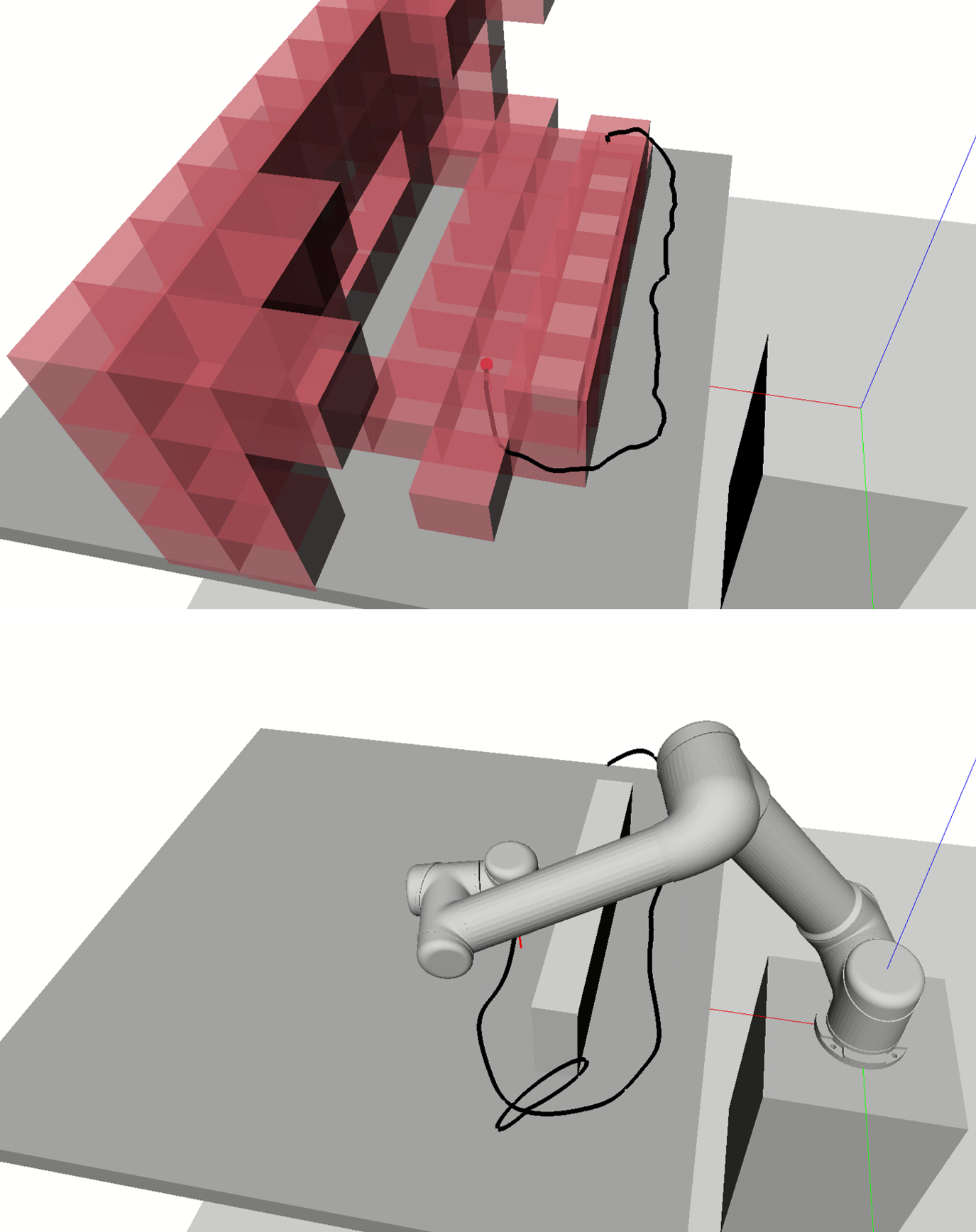}
         \caption{1st segment of unsuccessful demonstration}
         \label{fig:sequence_unguid_1}
     \end{subfigure}
     \begin{subfigure}[b]{0.24\textwidth}
         \centering
         \includegraphics[width=\textwidth]{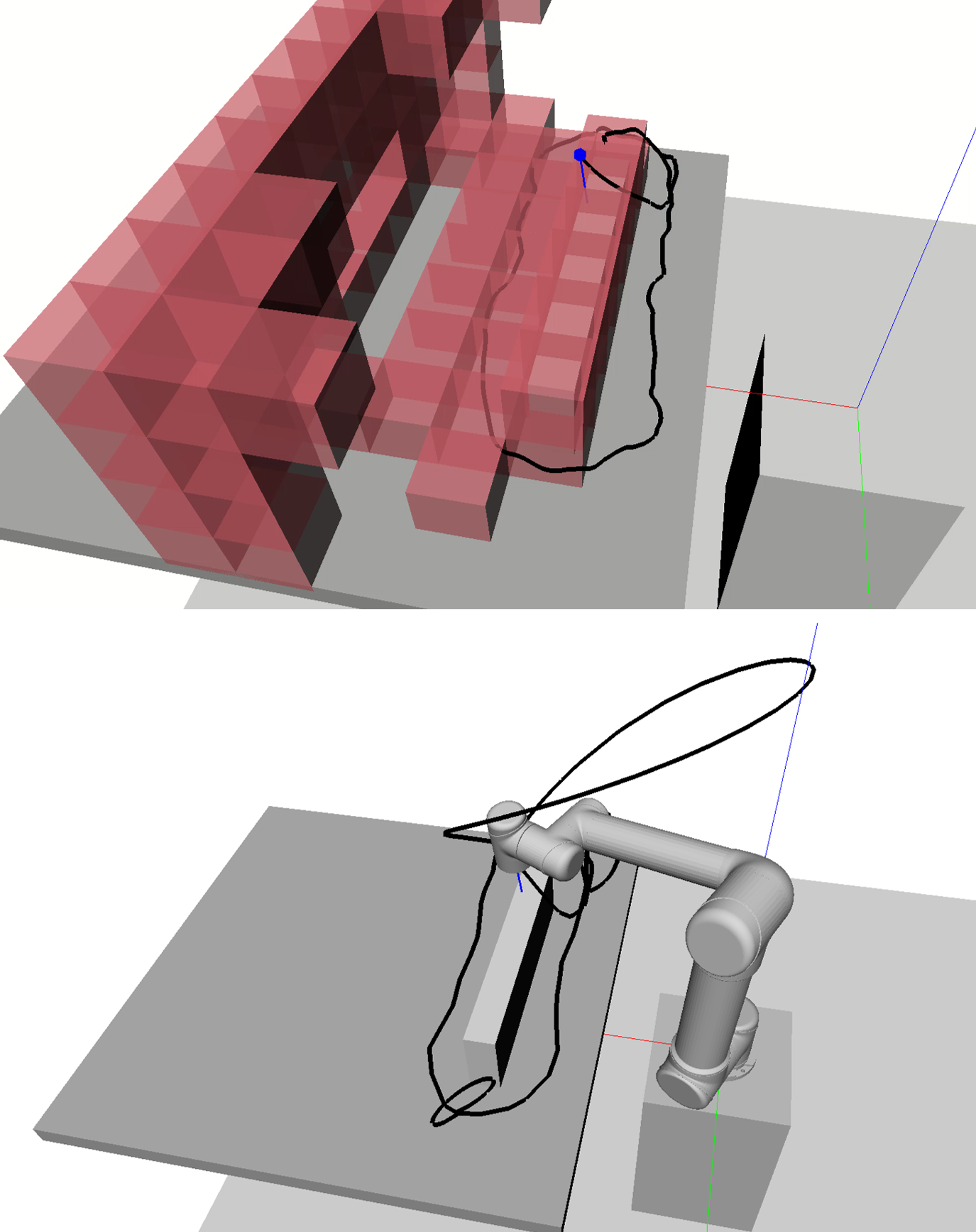}
         \caption{2nd segment of unsuccessful demonstration}
         \label{fig:sequence_unguid_2}
     \end{subfigure}
     \begin{subfigure}[b]{0.24\textwidth}
         \centering
         \includegraphics[width=\textwidth]{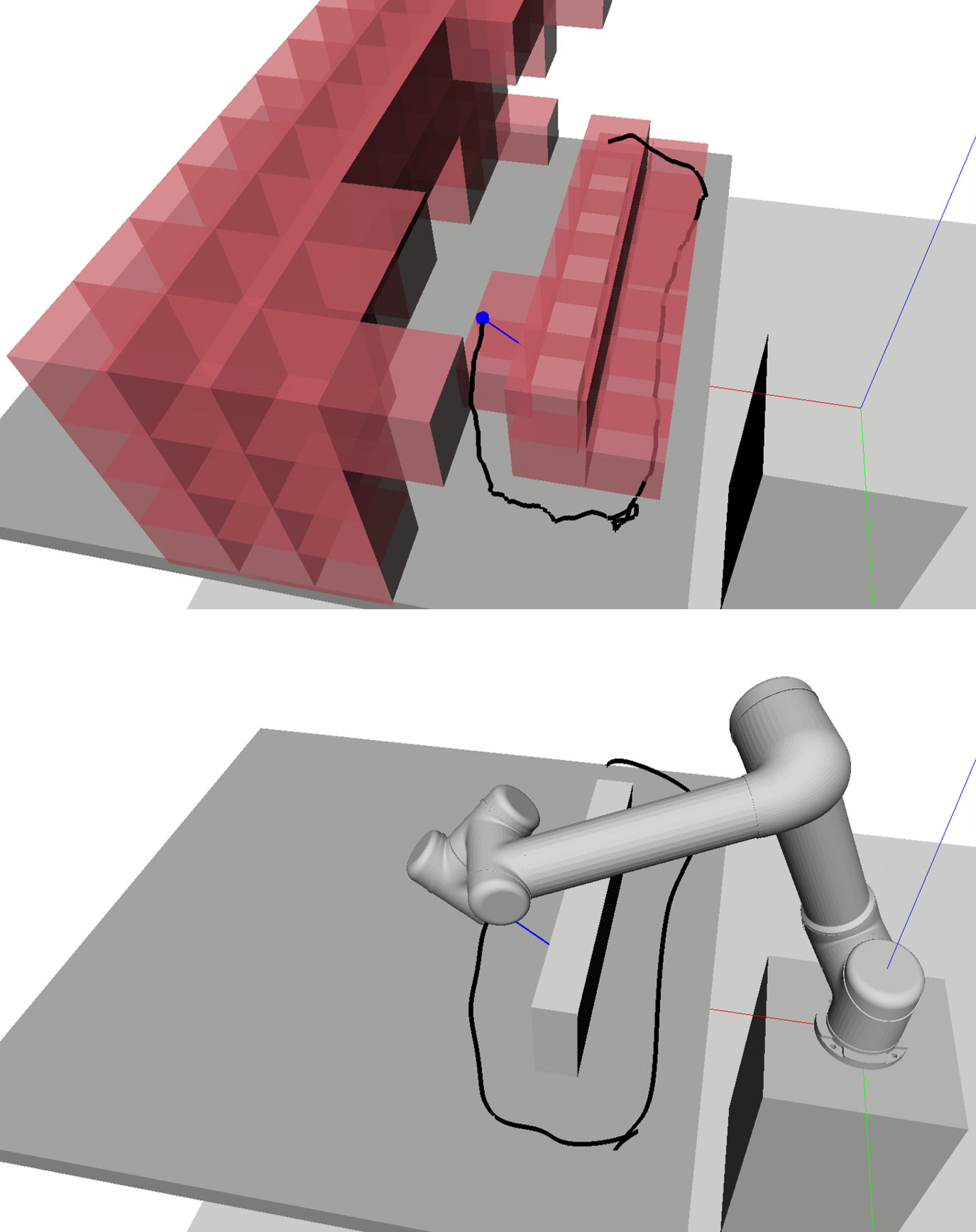}
         \caption{1st segment of successful demonstration}
         \label{fig:sequence_guid_1}
     \end{subfigure}
      \begin{subfigure}[b]{0.24\textwidth}
         \centering
         \includegraphics[width=\textwidth]{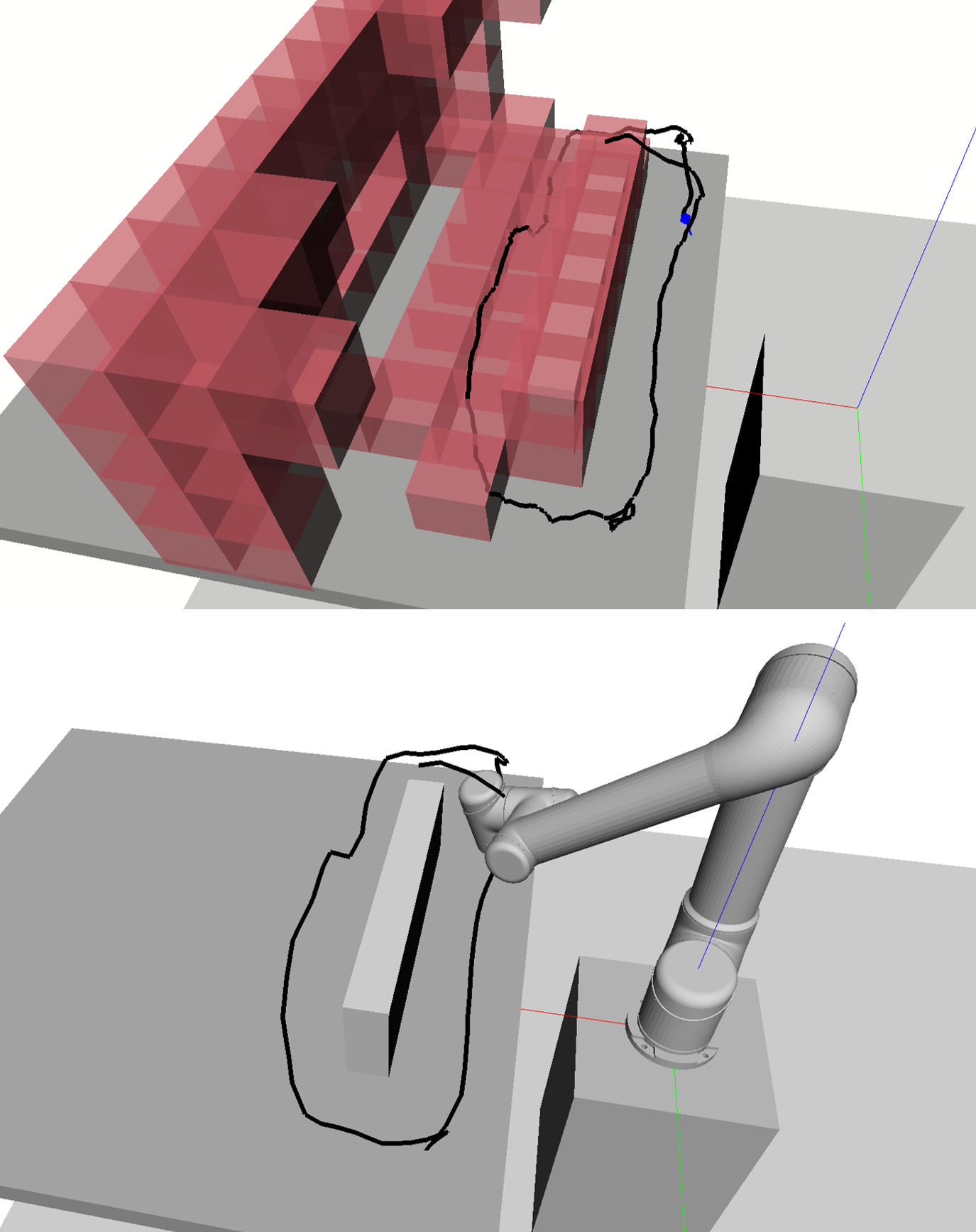}
         \caption{2nd segment of successful demonstration}
         \label{fig:sequence_guid_2}
     \end{subfigure}
        \caption{Two example demonstration sequences with GUI overlaid (top row) and their corresponding reproduced motion (bottom row). The sequence in (\subref{fig:sequence_unguid_1})-(\subref{fig:sequence_unguid_2}) attempts a downward facing orientation of the end effector on the far side of the wall and fails due to running into the red voxels in the GUI. The sequence in (\subref{fig:sequence_guid_1})-(\subref{fig:sequence_guid_2}) succeeds due to pitching back the orientation of the end effector.}
        \label{fig:showcase_experiment}
\end{figure*}

\subsection{Task Model}
As $SE(3)$ is the common representation space, $\lambda_{dem}$ and $\lambda_{tar}$ are a sequence of poses represented as position, $\mathbf{p}$, and quaternion, $\mathbf{q}$. We choose Cartesian space Dynamic Movement Primitives (CDMPs) for the encoding of the task as they are considered a prominent robot-independent task model in learning from demonstration. However, the general idea can be extended to other trajectory-based LfD task models.

A single demonstration of a desired motion is recorded as a trajectory $\lambda_{dem} = \{t_i,\mathbf{p}_i,\mathbf{q}_i\}\big|^T_{i=0}$. The task model consists of two parameterised dynamical second-order systems
with an additional term for external forces which shapes the trajectory to match the demonstration.
The nonlinear shape of the demonstrated trajectory is modelled via weighted kernel functions, e.g., radial basis functions. For producing $\lambda_{tar}$, the dynamical systems including external wrenches with learned weights are numerically solved, resulting in a continuous
model representation in $SE(3)$. In our work, we applied the revised bio-inspired formulation by Koutras et al.~\cite{Koutras2019, Koutras2020}. For an in-depth explanation of DMPs see \cite{Saveriano2021}.

\subsection{Motion Reproduction} 
For reproducing the demonstrated motion on $\mathcal{S}_{tar}$ we utilise the $\epsilon$-GHA mapping found by HAP to map $\lambda_{tar}$ to a sequence of arm configurations, $\pi_{tar} = \{ \theta_1, \theta_2, ... , \theta_T\}$. However, poses in $\lambda_{tar}$ are in continuous space where as the $\epsilon$-GHA mapping is defined over a discrete task space. Hence, in order to remain within $\mathcal{R}*$ a configuration is assigned to each $\theta \in \pi_{tar}$ that is close to the $\epsilon$-GHA mapping. We achieve this by computing all the IK solutions for each $x \in \lambda_{tar}$ and then choosing the the IK solution that minimises the euclidean distance to any of the $k$ closest mapped configuration in $\mathcal{R}*$.

\section{EXPERIMENTS}

\begin{figure*}[h]
     \centering
     \begin{subfigure}[b]{0.24\textwidth}
         \centering
         \includegraphics[width=\columnwidth]{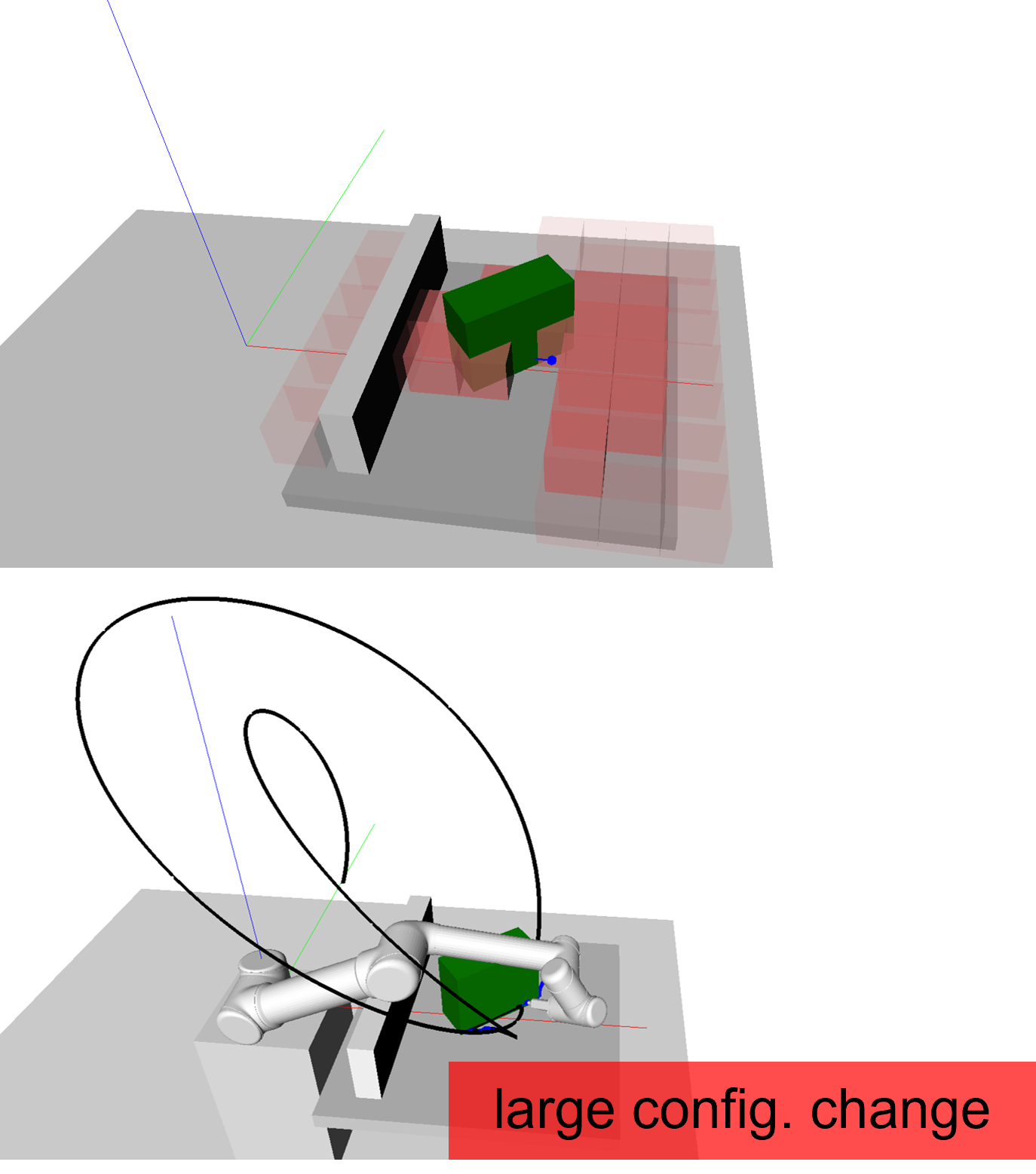}
         \caption{Failure due to large configuration change (non-guided).}
         \label{fig:failure1_large_config_change}
     \end{subfigure}
     \begin{subfigure}[b]{0.24\textwidth}
         \centering
         \includegraphics[width=\columnwidth]{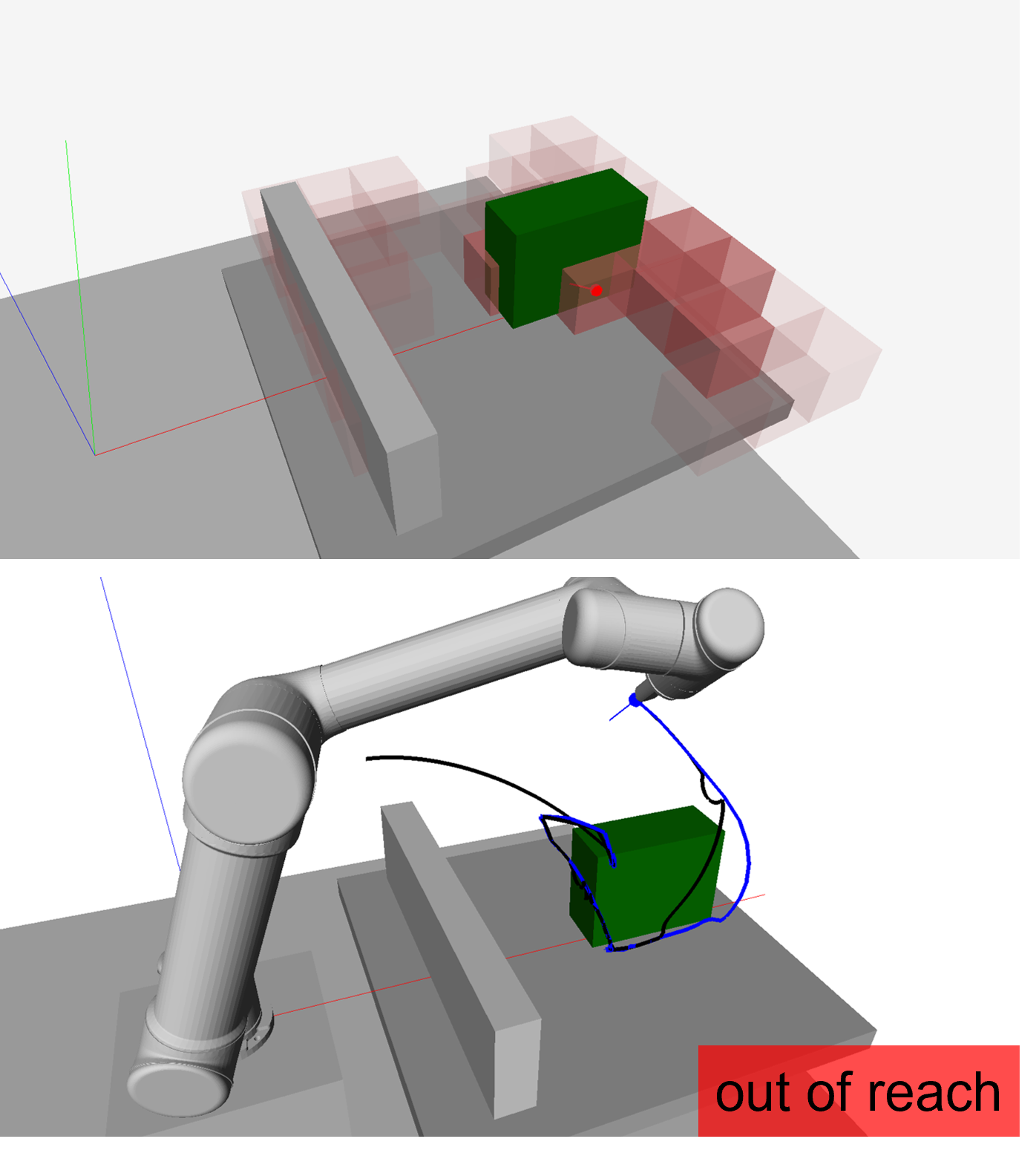}
         \caption{Failure due to unreachable weld path (non-guided).}
         \label{fig:failure2_out_of_reach}
     \end{subfigure}
     \begin{subfigure}[b]{0.24\textwidth}
         \centering
         \includegraphics[width=\columnwidth]{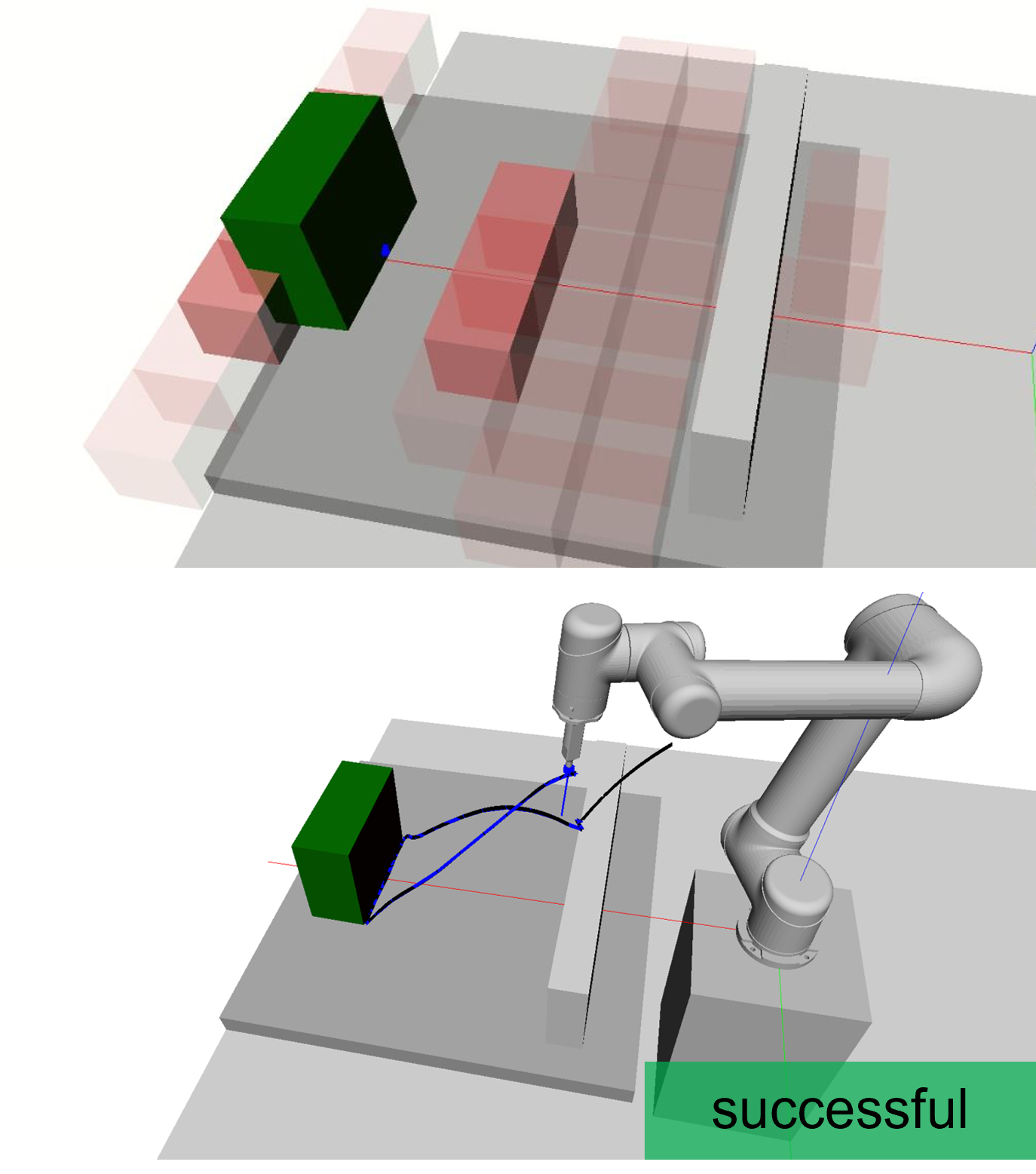}
         \caption{Successful object placement 1 (guided).}
         \label{fig:success1_far_away}
     \end{subfigure}
      \begin{subfigure}[b]{0.24\textwidth}
         \centering
         \includegraphics[width=\columnwidth]{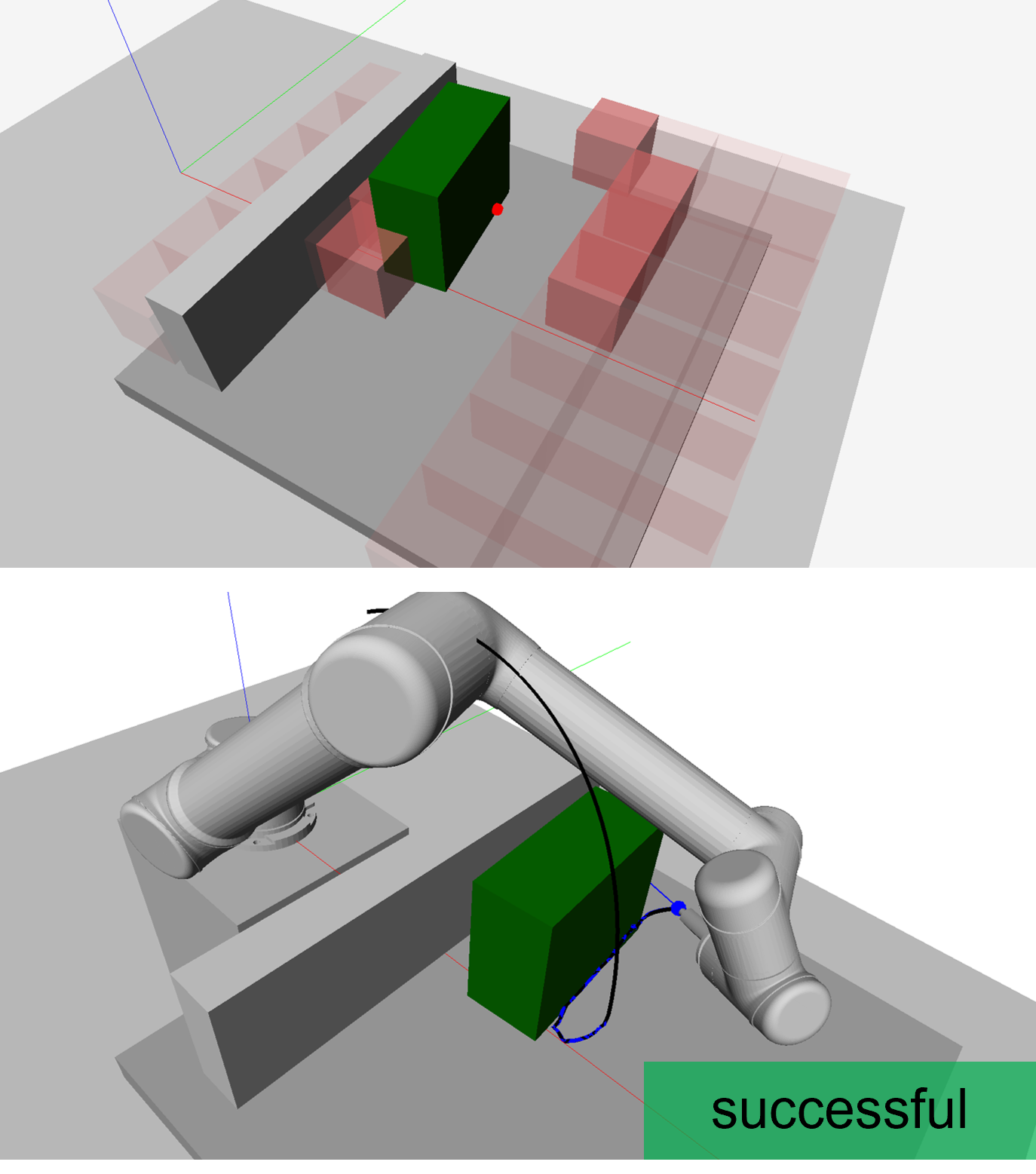}
         \caption{Successful object placement 2 (guided).}
         \label{fig:success2_close}
     \end{subfigure}
        \caption{Representative user study results showing diverse success and failure cases. Reproduced motions (bottom row) and corresponding GUI output (top row) are shown.  (\subref{fig:failure1_large_config_change}) and (\subref{fig:failure2_out_of_reach}) are two non-guided demonstrations that failed (GUI shown here for convenience - not shown to users). (\subref{fig:success1_far_away}) and (\subref{fig:success2_close}) are two successfully demonstrated tasks using our guided method.}
        \label{fig:success_failure_examples}
\end{figure*}

\begin{figure}[ht]
    \centering
     \begin{subfigure}[b]{0.48\columnwidth}
         \centering
         \includegraphics[width=0.95\columnwidth]{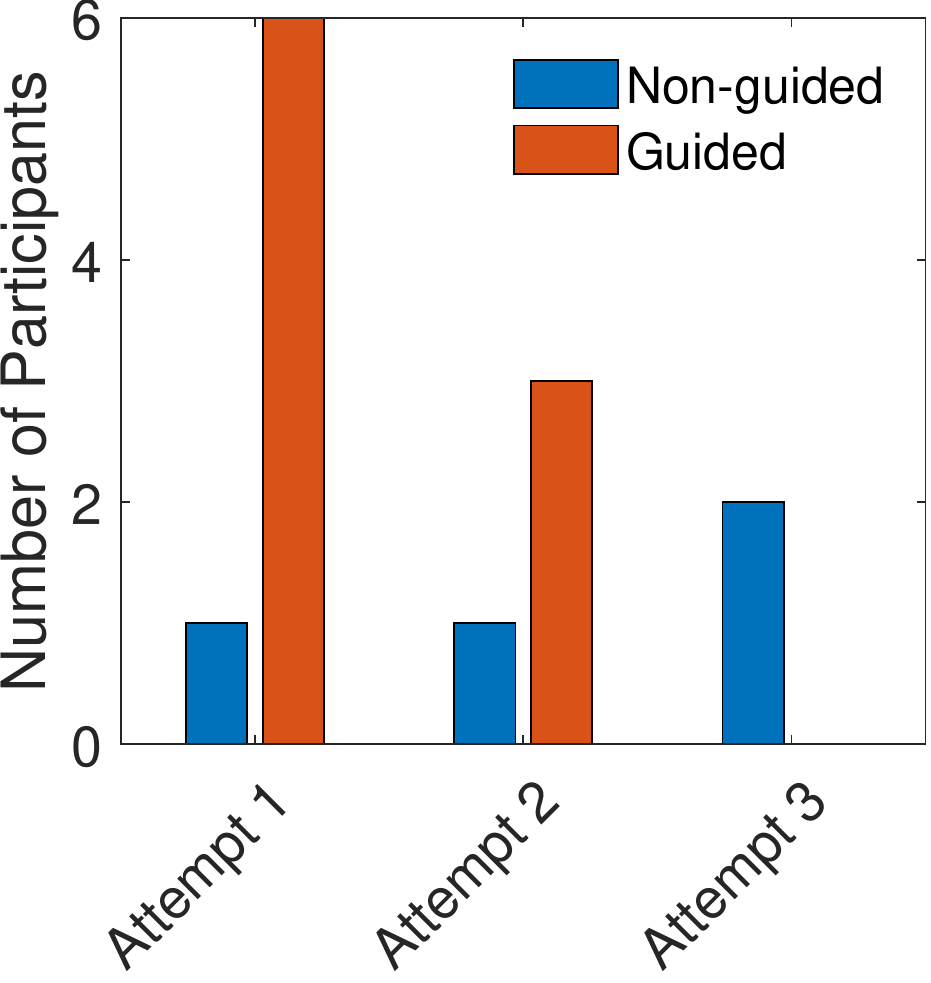}
         \caption{Distribution of number of trials required for success}
         \label{fig:stats_rep_attempts}
     \end{subfigure}
     \begin{subfigure}[b]{0.48\columnwidth}
         \centering
         \includegraphics[width=\columnwidth]{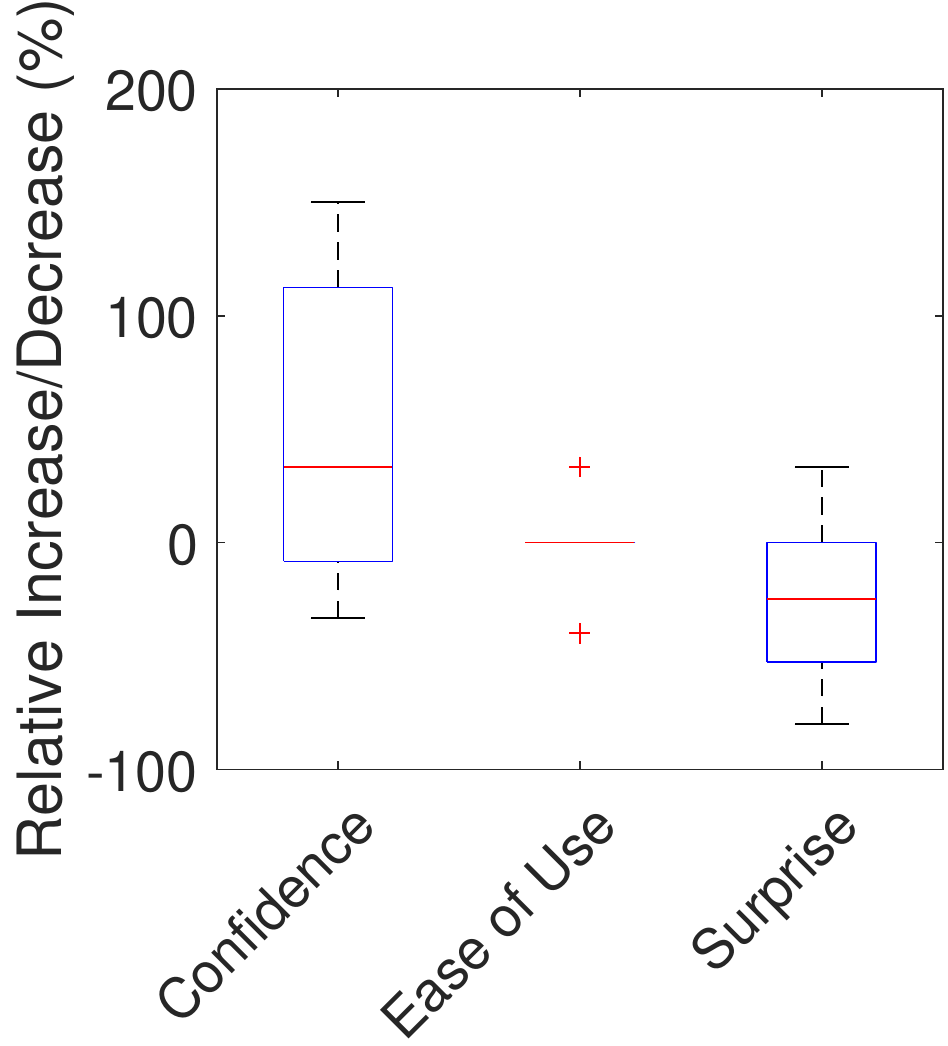}
         \caption{Percentage increase/ decrease in subjective metrics}
         \label{fig:stats_comparison}
     \end{subfigure}
    \caption{Quantitative results from the user study. (\subref{fig:stats_rep_attempts}) shows the distribution of number of trials required to successfully demonstrate the weld task and
    (\subref{fig:stats_comparison}) shows the percentage increase/ decrease in the collected subjective metrics.}
    \label{fig:userstudy_stats}
\end{figure}

This section presents the experimental evaluation of the proposed guided learning from demonstration approach. A validation experiment and a comprehensive user study were conducted to showcase the efficacy and usability of our guided learning from demonstration method.
In both experiments the interactive GUI is displayed to the demonstrator on a monitor during demonstration. The environment and task space discretisation strategy used in the experiments is the same as shown in Fig.~\ref{fig:example_lfd_scenario}. Additional footage from the experiments is included in the attached supplementary video.

\subsection{Validation}
In order to validate our approach we utilise a redundant seven-DOF Sawyer arm as the demonstration system, which is moved kinaesthetically (see Fig.~\ref{fig:exp_sawyer}). The target system is a simulated six-DOF UR5. In this experiment we demonstrate motions around the wall obstacle and aim to show that avoiding red voxel regions displayed by the GUI leads to successful reproduction and vice versa when not.

Example demonstration sequences and their reproduced motions are shown in Fig.~\ref{fig:showcase_experiment}. The sequence in Figs.~\ref{fig:sequence_unguid_1}~and~\ref{fig:sequence_unguid_2} attempts to move around the wall in an orientation normal to the workbench for the entire trajectory. However, as can be seen in the corresponding GUI snapshots this leads to moving through the red voxels. The explanation for this is that the arm must perform a large joint angle change in order to avoid colliding its wrist with the wall. This results in a poorly reproduced end effector trajectory, shown as a black line in the bottom row of Fig.~\ref{fig:sequence_unguid_2}.

In contrast, the demonstrated sequence in Figs.~\ref{fig:sequence_guid_1} and~\ref{fig:sequence_guid_2} avoids this problem by pitching back the orientation of the end effector. In this orientation the red voxels on the far side of the wall disappear and the trajectory is able to be reproduced successfully. This observed behaviour validates our method for this particular scenario and confirms our claim that reproduction can fail if the kinematics of the target system are not considered when guiding demonstrations, even in a relatively simple environment.

\subsection{User Study}

A subsequent user study was conducted with nine participants in order to evaluate the usability of our system. Users were tasked with carrying out a mock weld demonstration which consisted of placing an object to be welded on the workbench, shown in Fig.~\ref{fig:frontpage} and then performing a weld along a marked edge of the object using a hand-held 3D printed tool tracked by a VICON system. Nine users participated in the study consisting of three women and six men, five of which had little to no experience with robot arms.

For the user study we additionally enhanced the GUI to assist users with depth perception by dynamically increasing the opacity of the voxels within some distance threshold of the weld tool position. Furthermore, red boxes above the position of the tool are temporarily removed to prevent obstructing the view of the demonstrator. An example of this is shown in Fig.~\ref{fig:frontpage} for a particular position and orientation of the weld tool. 

Regarding evaluation against other methods, existing work that deals with transferring motion between two different systems utilise null-space control. Since the target system is non-redundant, the former is not possible.

Thus, first each user was asked to perform the task without the guidance of the interactive GUI. They were given a maximum of three attempts which involved carrying out the demonstration and then verifying that it was reproducible on a simulated UR5 in the same environment. If successful the reproduction was carried out on the real UR5 robot.

Afterwards, the user was asked to carry out the same task with the guidance of the GUI, again with three attempts. If the user was successful in the unguided attempt they were asked to find another valid placement of the weld object. The main advantage of having the GUI was that it helped users make an informed decision on where to place the weld object. Furthermore, users were able to place the object and then update the region of feasible motion before carrying out the demonstration.

\begin{figure*}[t]
     \centering
     \begin{subfigure}[b]{0.19\textwidth}
         \centering
         \includegraphics[width=\textwidth]{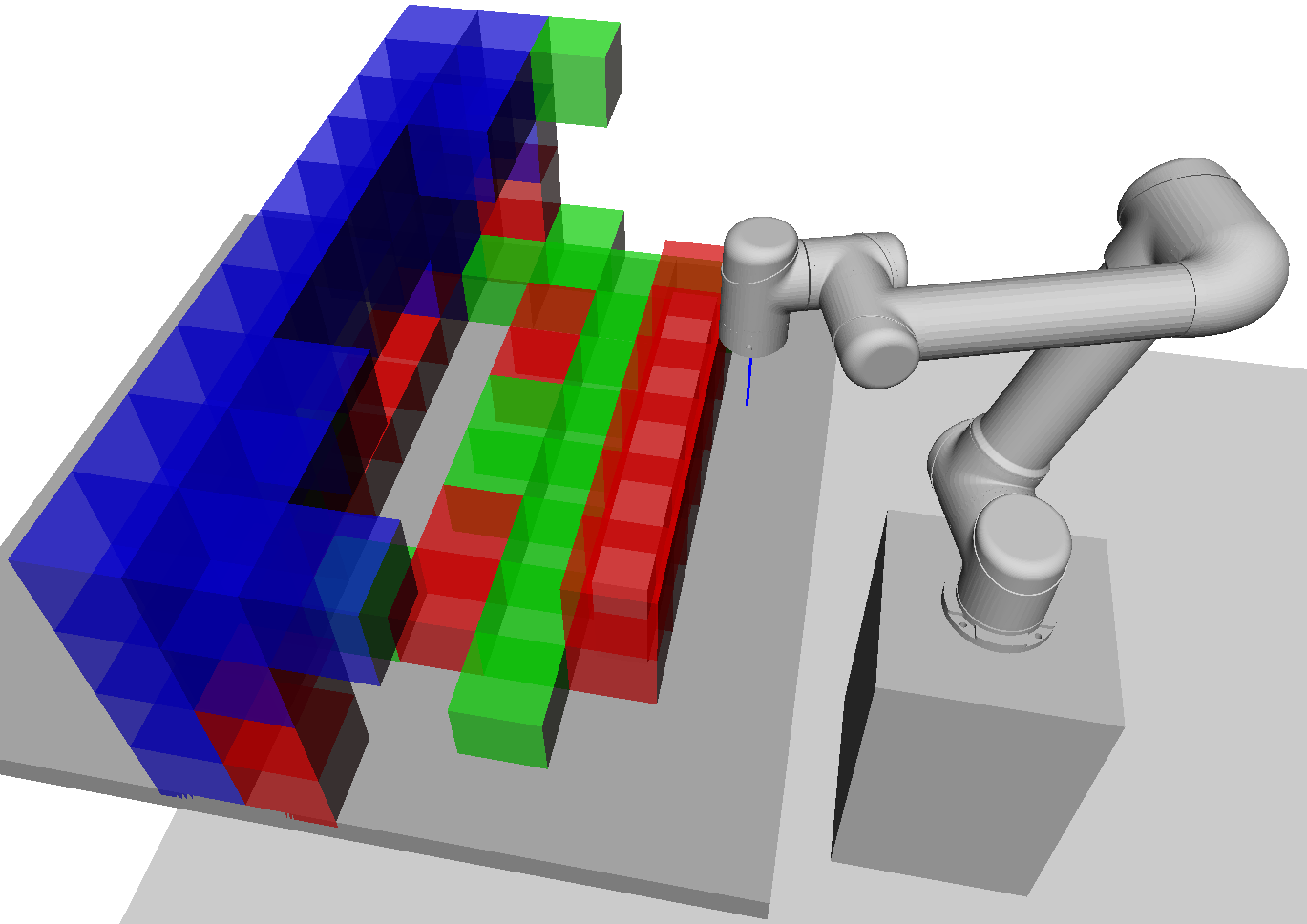}
         \caption{Orientation 1}
         \label{fig:class_orientation1}
     \end{subfigure}
     \begin{subfigure}[b]{0.19\textwidth}
         \centering
         \includegraphics[width=\textwidth]{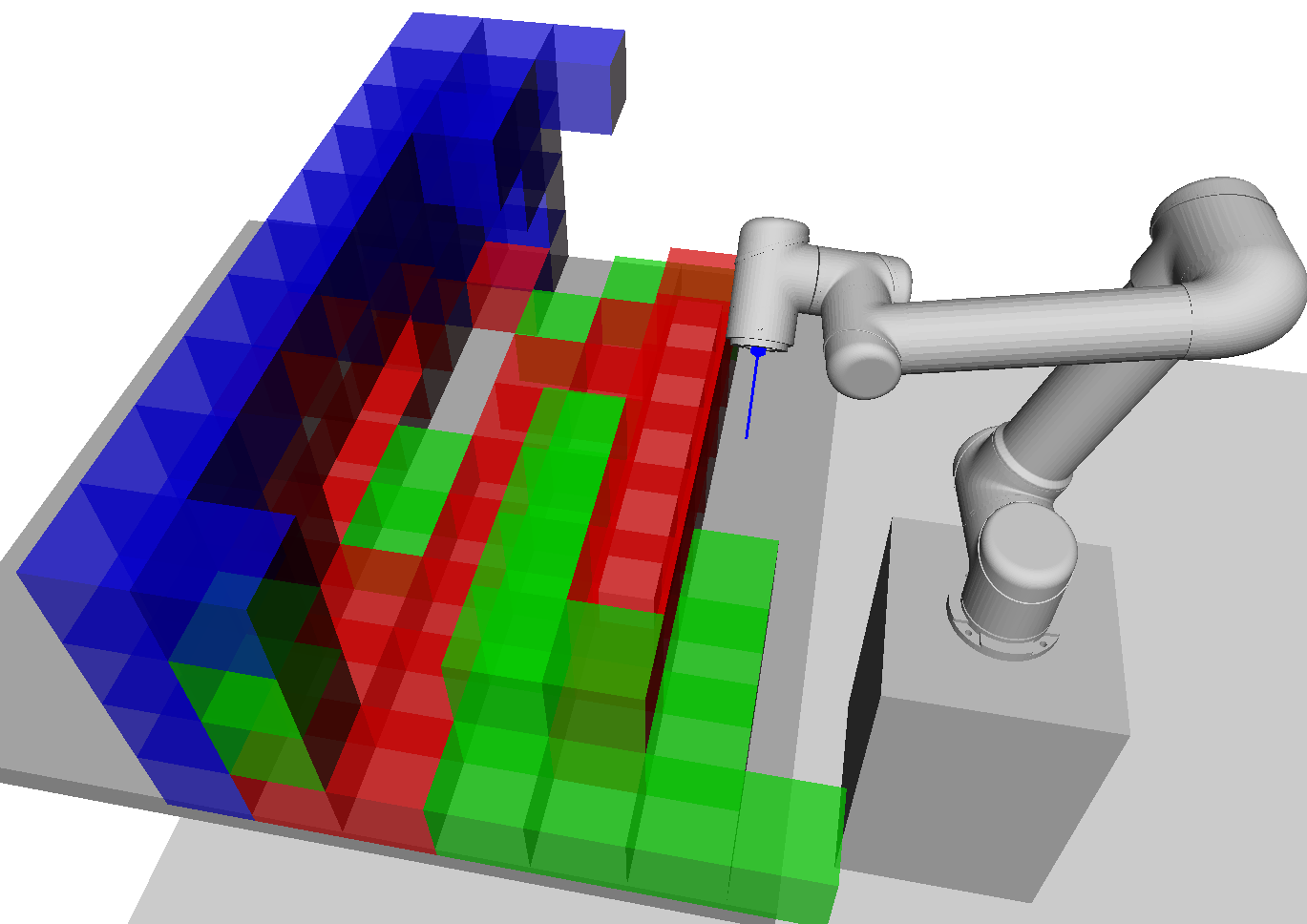}
         \caption{Orientation 2}
         \label{fig:class_orientation2}
     \end{subfigure}
      \begin{subfigure}[b]{0.19\textwidth}
         \centering
         \includegraphics[width=\textwidth]{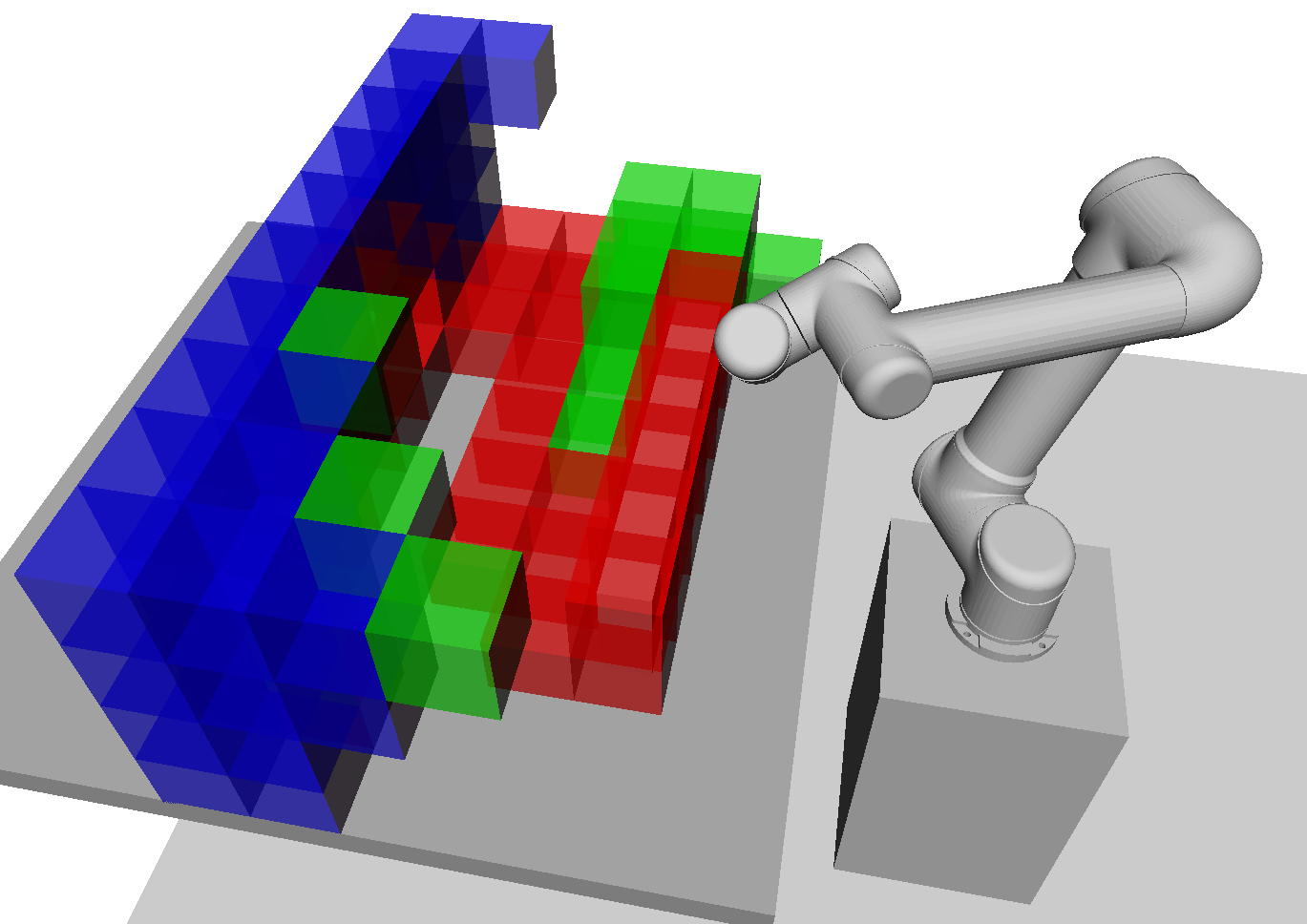}
         \caption{Orientation 3}
         \label{fig:class_orientation3}
     \end{subfigure}
      \begin{subfigure}[b]{0.19\textwidth}
         \centering
         \includegraphics[width=\textwidth]{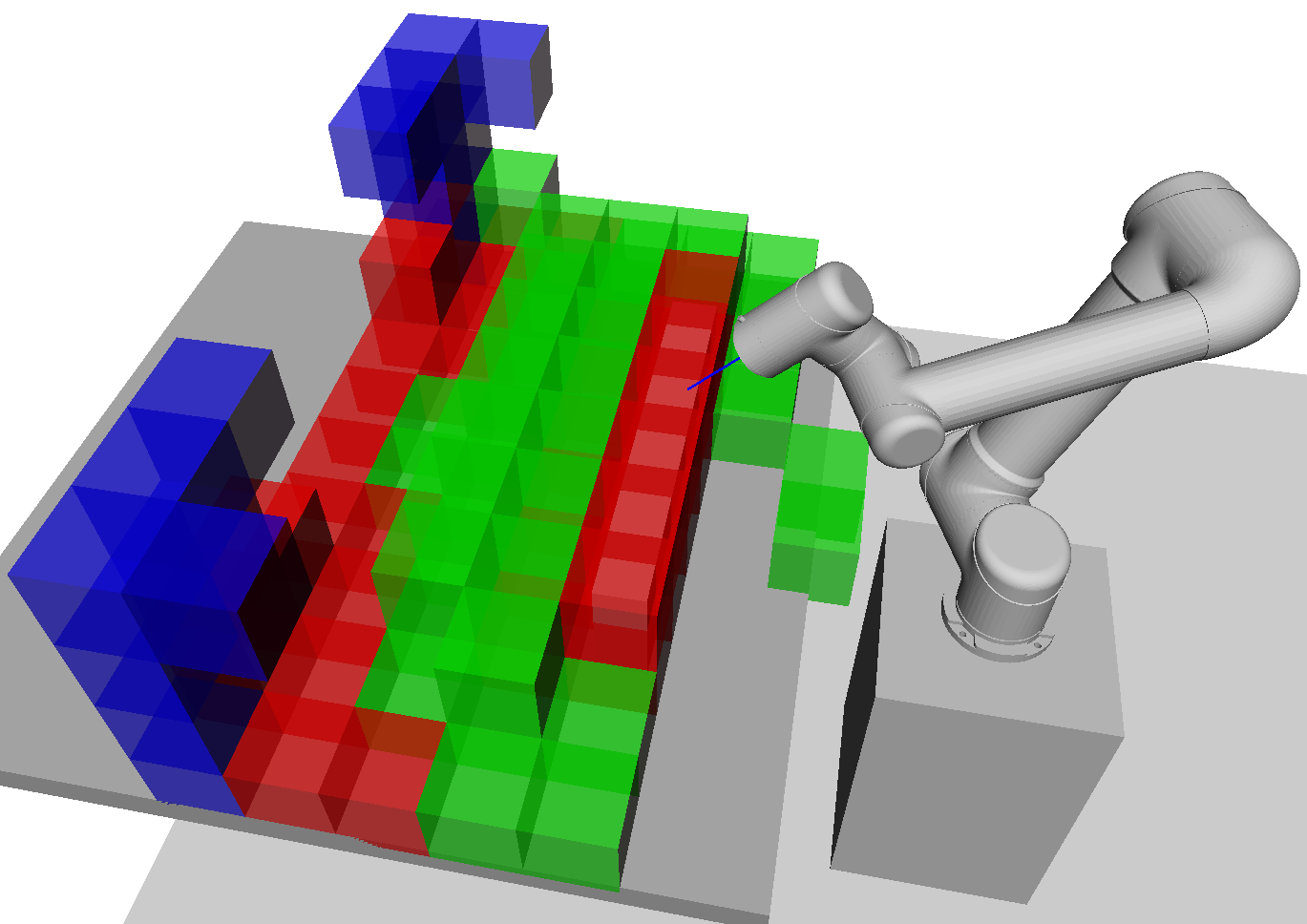}
         \caption{Orientation 4}
         \label{fig:class_orientation4}
     \end{subfigure}
     \begin{subfigure}[b]{0.19\textwidth}
         \centering
         \includegraphics[width=\textwidth]{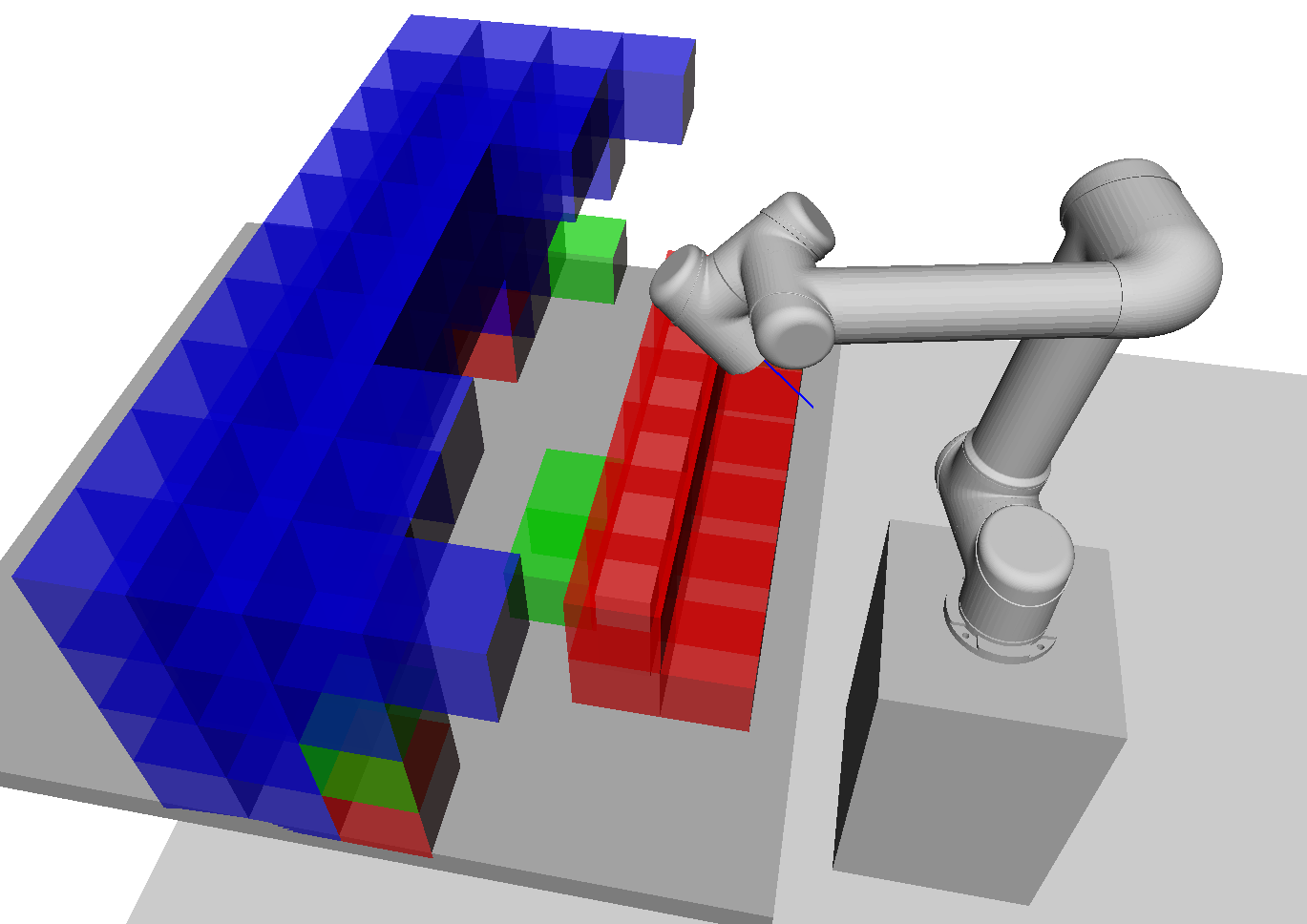}
         \caption{Orientation 5}
         \label{fig:class_orientation5}
     \end{subfigure}
        \caption{Different regions of the subspace being blocked due to: All IK solutions in collision (red), requiring large change in configuration in order to reach (green) and no IK solution due being outside the arm's reachability (blue).}
        \label{fig:voxel_classes}
\end{figure*}

Representative trials from the user study are shown in Fig.~\ref{fig:success_failure_examples}. Quantitative results are summarised in Fig.~\ref{fig:userstudy_stats}. Fig.~\ref{fig:stats_rep_attempts} shows the distribution of the number of trials required to successfully demonstrate the weld task. Note that five out of the nine participants were unsuccessful in all three attempts for the non-guided experiment. As can be seen, all users were able to successfully demonstrate the task within two attempts with guidance, where as a majority of users - 56\% - failed to do so at all without guidance.

To assess usability we chose subjective metrics based on a previous usability study on socially assistive robots~\cite{olde2019using}. Their study categorised their evaluation metrics into effectiveness, efficiency and satisfaction with which users were able to achieve given tasks~\cite{iso20189241}. In a similar fashion we asked users to rate on a 5-point scale how confident they were and how easy it was to carry out the task with and without the GUI. In addition, we asked how surprised they were at the reproduced trajectory. The percentage increase and decrease in each of these metrics is shown in Fig.~\ref{fig:stats_comparison}.

As can be seen, almost all participants were more confident when carrying out the task when utilising guidance from the GUI. Ease of use was comparable which is a positive result considering the addition of the GUI during demonstration. And users were less surprised by the reproduced motions when using the GUI which can be explained by the absence of large, spontaneous joint changes.

We additionally asked participants the following questions on a 5-point scale specifically about the effectiveness and design of the GUI: 
\begin{itemize}
  \item How effective was the GUI at improving your decision of where to place the weld object?
  \item How intuitive was the GUI in assisting with the given task?
  \item How effective was the GUI in providing spatial awareness, i.e. avoid red voxels?
  \item Is the extra effort to use the GUI worth it over trial and error (without any guidance)?
\end{itemize}
On average users gave a rating of 4 and above which indicates that our GUI design was overall effective and intuitive enough for assisting with the given task. Of particular note was the fact that all participants rated the last question a 5 which reinforces the value of our method over a trial and error approach.

In terms of qualitative feeedback, some participants desired more suggestive guidance. For example providing recommended object placements to the user. Other suggested improvements included
receiving more information on the source of failure when entering a red voxel and the option to display the simulated robot in the GUI.

\section{Discussion}

\subsection{Analysing $\bar{\mathcal{R}*}$}
To explain the poses in $\bar{\mathcal{R}*}$, color coded voxels are shown in Fig.~\ref{fig:voxel_classes} for varying end-effector orientations. The green voxels behind the wall are blocked due requiring a large change about the wrist in order to avoid collision with the wall or table which would violate our region of reproducible motions condition. The red voxels indicate regions where no IK solution is possible without colliding with the environment. The blue voxels indicate regions outside the reachability of $\mathcal{S}_{tar}$.

\subsection{Practical Considerations}
We designed the discretised task space for the experiments, shown in Fig.~\ref{fig:example_lfd_scenario}, such that all poses were approximately oriented in a nominal direction, facing down into the workbench. Allowing any greater deviations from this nominal direction would result in an ambiguous guidance in the GUI. For example, it may be possible to rotate the end effector about a point in the workspace in one direction but rotating in the other direction may require a large joint angle change to avoid self-collision or joint limits. However, this is a limitation of the GUI rather than our method for finding $\mathcal{R}*$.
A work around is to define multiple task spaces with the required nominal directions and then carry out demonstrations independently for each of these.

Furthermore, the $\mathcal{R}*$ found is approximated by this discretisation. Hence if more granularity is required for more fine motor tasks one could increase the resolution. To combat the increase in computation, one could potentially employ a multi-resolution strategy.

\section{CONCLUSIONS}
We presented a new guided learning from demonstration formulation for ensuring reproducibility when transferring a demonstrated motion to a system with differing kinematic structure. To address this problem we leveraged an existing robotic manipulator planner to compute regions of reproducible motions and then utilised this to provide intuitive guidance to the demonstrator via a novel interactive GUI. Results from our user study showcased the significance of being able to demonstrate reproducible motions reliably without the presence of the physical target system.

Future work includes improving the GUI design to allow for more flexible operational space definition and exploring other guidance mediums such as augmented reality and haptic feedback. Presenting suggestions to user, such as where to place objects in order to carry out a task successfully, is another important avenue for improving usability.

\balance
\bibliographystyle{IEEEtran}
\bibliography{references}

\end{document}